\documentclass[runningheads]{llncs}
\usepackage[T1]{fontenc}
\usepackage{graphicx}
\usepackage{enumitem}
\usepackage{amssymb}
\usepackage{xcolor}
\usepackage[normalem]{ulem}
\usepackage{amsfonts}
\usepackage{ulem}
\usepackage{amsmath, amssymb}

\begin{document}

\title{TLRN: Temporal Latent Residual Networks For Large Deformation Image Registration}

\author{
Nian Wu\inst{1}  
\and
Jiarui Xing\inst{1} 
\and
Miaomiao Zhang\inst{1,2} 
}

\institute{Department of Electrical and Computer Engineering, University of Virginia, USA \and
Department of Computer Science, University of Virginia, USA
}

\maketitle              
\begin{abstract}
This paper presents a novel approach, termed {\em Temporal Latent Residual Network (TLRN)}, to predict a sequence of deformation fields in time-series image registration. The challenge of registering time-series images often lies in the occurrence of large motions, especially when images differ significantly from a reference (e.g., the start of a cardiac cycle compared to the peak stretching phase). To achieve accurate and robust registration results, we leverage the nature of motion continuity and exploit the temporal smoothness in consecutive image frames. Our proposed TLRN highlights a temporal residual network with residual blocks carefully designed in latent deformation spaces, which are parameterized by time-sequential initial velocity fields. We treat a sequence of residual blocks over time as a dynamic training system, where each block is designed to learn the residual function between desired deformation features and current input accumulated from previous time frames. We validate the effectivenss of TLRN on both synthetic data and real-world cine cardiac magnetic resonance (CMR) image videos. Our experimental results shows that TLRN is able to achieve substantially improved registration accuracy compared to the state-of-the-art. Our code is publicly available at https://github.com/nellie689/TLRN.

\end{abstract}
\section{Introduction}
\label{sec:intro}
Temporal/time-series image registration plays an important role in various domains, including medical imaging~\cite{singh2006image,liao2016temporal}, video analysis~\cite{ghanem2012robust}, and remote sensing~\cite{yang2018multi}. This process involves aligning images captured at a sequence of time points; hence enabling the analysis and tracking of dynamic changing process\allowbreak\cite{morais2013cardiac,perperidis2005spatio,xing2024multimodal,geng2009implicit}. In the domain of medical applications, it is desirable that the estimated deformations being diffeomorphisms (i.e., one-to-one smooth and invertible smooth mappings between images~\cite{beg2005computing}). Such properties are essential to preserve the local and global topological structure of studied subjects, ensuring biologically meaningful results~\cite{hong2017fast,niethammer2011geodesic}.

A major challenge of temporal image registration is the large deformation propagated over time, for example, the start of a cardiac cycle compared to the peak stretching phase in CMR videos~\cite{xing2024multimodal,morais2013cardiac,perperidis2005spatio}. Traditional pairwise registration methods, which align images to a single reference image, often suffer from accumulating errors over time and fail to capture the complex and continuous nature of large deformations~\cite{reinhardt2008registration,chen2022transmorph,hinkle2018diffeomorphic,dalca2019unsupervised}. To alleviate this issue, existing approaches have been proposed to concatenate transformations estimated from consecutive frames with relatively small deformations~\cite{reinhardt2008registration,csapo2007image}. However, these methods often overlook the temporal dynamics of deformation fields or motion in the image data, resulting in compromised registration accuracy. Recognizing this limitation, several research focused on leveraging temporal information to enforce smoothness and continuity over time~\cite{liao2016temporal}. Other related studies introduced spatio-temporal parametric models based on B-splines to enforce the temporal consistency of estimated deformation fields~\cite{ledesma2005spatio,metz2011nonrigid,de2012temporal}. With the advancement of deep learning, predictive registration methods have gained popularity due to their faster inference compared to traditional optimization-based approaches~\cite{dalca2019unsupervised,chen2022transmorph,wang2020deepflash,hinkle2018diffeomorphic}. Recent works have harnessed the power of deep networks to generate deformation fields from a sequence of image videos while considering general temporal smoothness~\cite{krebs2021learning,krebs2020probabilistic,qiao2023cheart,qin2023generative}. Despite achieving impressive results, these methods do not explicitly model the relationships between transformations propagated over time, resulting in limited capability of capturing the underlying representations of large and complex deformations in dynamic processes.

In this paper, we introduce a novel approach, termed TLRN, to effectively capture fine details and complex patterns of large deformations by incrementally refining learned latent deformation features across time. To achieve this, we carefully design a multi-level structure of residual blocks in the temporal latent velocity space to model the hierarchical relationships between deformations. Each level builds upon and refines the feature pace learned by the previous levels. The main contributions of our proposed method are twofold:
\begin{enumerate}[label=(\roman*)]
\item Develop a temporal residual network in the latent space of velocity fields to explicitly model the hierarchical relationship between time-sequential deformations. 
\item Effectively incorporate the temporal smoothness and continuity of deformations/motions inherent in image videos, leading to improved registration accuracy. 
\end{enumerate}
We validate the proposed model, TRNL, using both synthetic data and real CMR image videos. Experimental results demonstrate that TRNL achieves superior registration accuracy, producing more robust and better regularized transformation fields compared to state-of-the-art deep learning-based registration networks.~\cite{dalca2019unsupervised,hinkle2018diffeomorphic,chen2022transmorph,joshi2022diffeomorphic}.

\section{Background: Diffeomorphic Image Registration}
In this section, we briefly review diffeomorphic image registration in the context of stationary velocity fields (SVF)~\cite{vercauteren2008symmetric}. Note that our model is general to other registration models, such as large deformation diffeomorphic metric mapping~\cite{beg2005computing}.

Given a source image, $S^A$, and a target image, $S^B$, defined on a $d$-dimensional torus domain $\Omega = \mathbb{R}^d / \mathbb{Z}^d$ ($S^A(x), S^B(x):\Omega \rightarrow \mathbb{R}$). The problem of diffeomorphic image registration is typically formulated as an energy minimization over a time-dependent deformation fields, $\{\phi_t:t\in[0,1]\}$, i.e.,
\begin{equation}
\label{eq:TotalEnergy}
E(\phi_1) = \frac{1}{2\sigma^{2}}\, \text{Dist}(S^A \circ \phi_1^{-1}, S^B) + \text{Reg}(\phi_1).
\end{equation}
Here, $\circ$ denotes an interpolation operator that deforms a source image to match the target. The Dist(·,·) is a distance function that measures the dissimilarity between images weighted by a positive parameter $\sigma$, and Reg($\cdot$) is a regularization term to enforce the smoothness of transformation fields. In this paper, we use a commonly used sum-of-squared intensity differences ($L_2$-norm)~\cite{beg2005computing,wu2023neurepdiff} as the distance function.

In the setting of SVF~\cite{vercauteren2008symmetric}, the diffeomorphic transformation fields, $\phi_t$, is parameteriz ed by a constant velocity field $v$ over time, i.e.,
\begin{eqnarray}
\label{eq:velocity_stationary}
\frac{d\phi_t}{dt} = v(\phi_t),   \,\, \text{s.t.}  \,\, \, \phi_0 = x.
\end{eqnarray}
The solution of Eq.~\eqref{eq:velocity_stationary} is identified with a group exponential map, which is numerically computed through a scaling and squaring method~\cite{vercauteren2008symmetric}. For a simplified notation, we will drop the time index in following sections, i.e., $\phi_1 \overset{\Delta}{=} \phi$.

\section{Our Method: TLRN}
\begin{figure*}[!b]
\centering
\includegraphics[width=1.0\textwidth] {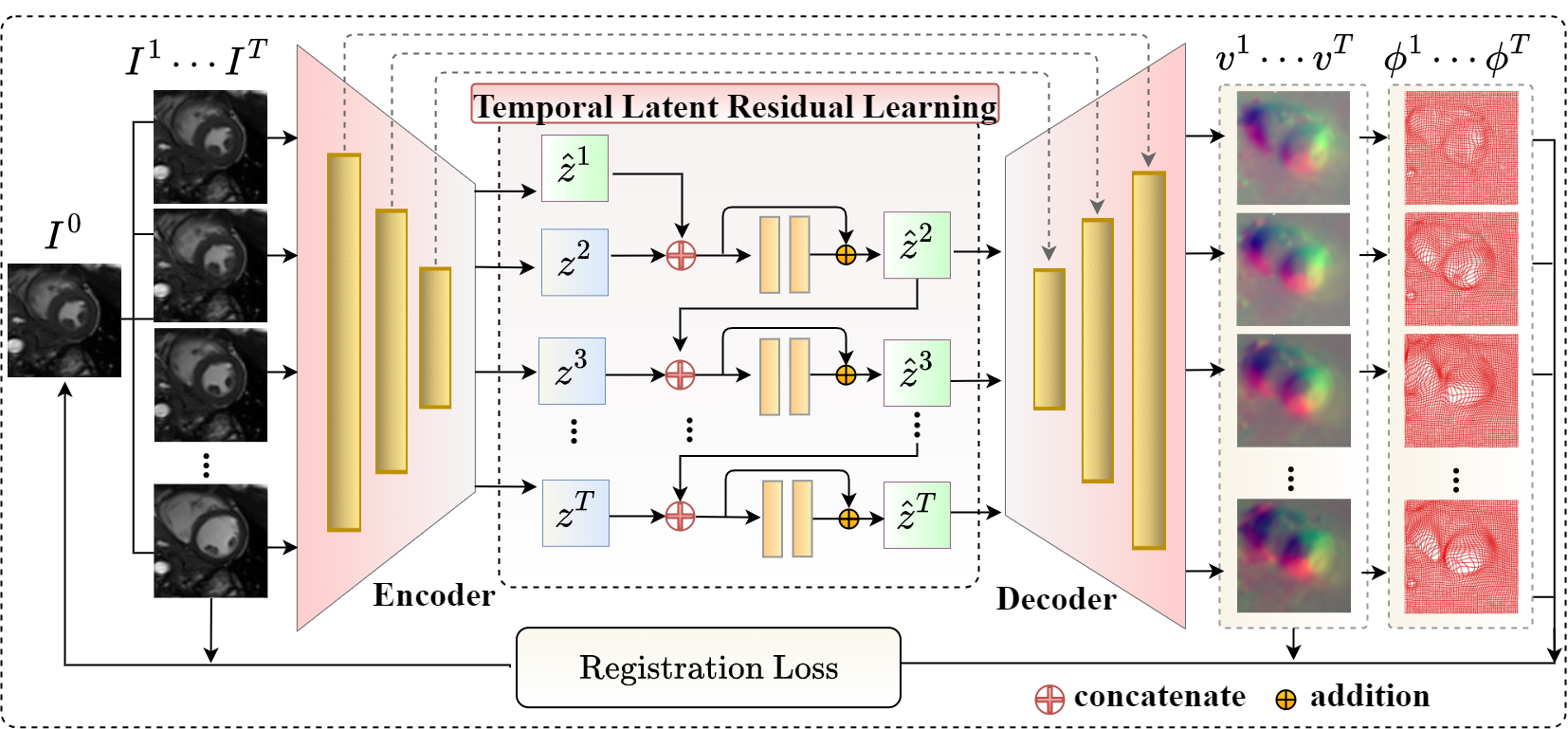}
     \caption{An overview of our proposed network TLRN.}
\label{fig:NetArch}
\end{figure*}

In this section, we introduce a novel temporal latent residual networks, TLRN, that can  effectively predict a sequence of deformation fields in time-series image registration. Our proposed TLRN consists of two key components: (i) an unsupervised registration network that learns the velocity fields for the input time-series images, and (ii) a temporal residual learning submodule designed in the learned latent velocity spaces to effectively adjust integrated deformation features from current and previous time steps. An overview of our proposed network is shown in Fig.~\ref{fig:NetArch}.

\subsection{Network Architecture}
\paragraph*{\bf Learning latent velocity fields via unsupervised registration.} Given a training dataset of $N$ image sequences, where each sequence includes $T+1$ time frames, denoted as $\{I^\tau_i\}, i \in \{1, \cdots, N\}, \tau \in \{0, \cdots, T\}$. That is to say, for the $i$th training data, we have a sequence of image frames $\{I^0_i, \cdots, I^T_i$\}. By setting the first frame $\{I^0_i\}$ as a reference (source) image, there exists a number of $T$ pairwise images, $\{(I_i^0, I_i^1), (I_i^0, I_i^2) \cdots, (I_i^0, I_i^T)\}$, to be aligned by their associated velocity fields $\{v_i^1, v_i^2 \cdots, v_i^T\}$. Similar to~\cite{dalca2019unsupervised,hinkle2018diffeomorphic}, we employ U-Net architecture as the backbone of our registration encoder, $\mathcal{E}_{\theta_v}$, and decoder, $\mathcal{D}_{\theta_v}$, parameterized by $\theta_v$. The encoder $\mathcal{E}_{\theta_v}$ projects the input image sequences into a latent velocity space $\mathcal{Z}$, which embeds a sequence of latent representations of the velocity fields, denoted as $\{ z^1, z^2, \cdots, z^T\} \in \mathcal{Z}$. The decoder $\mathcal{D}_{\theta_{v}}$ is then used to project the latent features back to the input image space. 

\paragraph*{\bf Temporal latent residual learning.} To leverage the temporal smoothness and continuity of time-series images, we develop a temporal residual learning scheme in the latent velocity space. Specifically, we introduce a latent recurrent unit with integrated residual blocks to perform two key tasks: (i) fusing the latent velocity features from current and previous time points, and (ii) re-adjusting these features through learned residual functions with reference to an optimal solution to further reduce error. Given the adjusted latent feature $\hat{z}^{\tau-1}$ at previous time point and $z^\tau$ at current time point, the output feature $\hat{z}^\tau$ of our proposed residual block can be represented as
\begin{align}\label{eq:RFU}
   \hat{z}^{\tau} := \mathcal{F}_{\theta_r}(\hat{z}^{\tau-1} \oplus z^{\tau}) + W_{\theta_s}(\hat{z}^{\tau-1} \oplus z^{\tau}),
\end{align}
where $\mathcal{F}_{\theta_r}$ represents a residual function parameterized by $\theta_r$ and $\oplus$ denotes vector concatenation. Following the principles in~\cite{he2016deep}, we perform a learnable linear projection, $W_{\theta_s}$, by the shortcut connections to match the dimensions of $\mathcal{F}_{\theta_r}$ and $(\hat{z}^{\tau-1} \oplus z^{\tau})$. Note that the choice of residual function, $\mathcal{F}_{\theta_r}$, is flexible. In this paper, we employ a composition of two convolutional layers and Leaky Rectified Linear Unit (LeakyReLU)~\cite{maas2013rectifier}.  
\paragraph*{\bf Network loss.} By defining $\Theta=(\theta_v, \theta_r, \theta_s)$ for all network parameters, we finally formulate the loss function of TLRN as
\begin{align}\label{eq:LossFun}
l(\Theta) =  \sum_{i=1}^{N} \sum_{\tau=1}^{T} \lambda \, \|(I_i^0 \circ \phi_i^{\tau} \left(v_i^{\tau} (\hat{z}^\tau_i;\Theta) \right)  - I^\tau_i \|^2_2 +  \| \nabla v_i^{\tau} (\hat{z}^\tau_i;\Theta) \|^2  + \text{reg} (\Theta).
\end{align}
Here, $\phi_i^{\tau}$ is the transformation fields between $(I_i^0, I_i^{\tau})$, $\lambda$ is a positive weighting parameter, and reg(·) represents network regularity. We jointly optimize all parameters till convergence. 

\section{Experimental Evaluation}
We validate our model, TLRN, on both synthetic data and real CMR image videos. All our experiments are trained on servers with AMD EPYC $7502$ CPU of $126$GB memory and Nvidia GTX $4090$Ti GPUs. We use the Adam optimizer~\cite{kingma2014adam} with $20000$ epochs, learning rate of $1e^{-4}$, and a batch size of $32$. 
\subsection{Datasets}
{\bf 2D Lemniscate Sequence.} We simulate a number of $1400$ "lemniscate" series ($64^2 \times T$), where $T + 1=12$ (examples are shown in Fig.~\ref{fig:synthetic}). For each reference image $I_i^0$, we first generate a 2D contour with its coordinates computed by $\{^{x = a \cdot \text{cos}(\alpha) / (\text{sin}^{2}(\alpha) + 1)}_{y = a \cdot \text{sin}(\alpha) \text{cos}(\alpha)/ (\text{sin}^{2}(\alpha) + 1)}$, where $a$ is a scaling factor, and $\alpha$ is uniformly sampled from $[0, 2 \pi]$. We then determine the contour thickness by varying the coordinates with parameters $\sigma_x$ and $\sigma_y$, i.e., $[x \pm \sigma_x, y \pm \sigma_y]$. The subsequent time frames $\{I_i^1, \cdots, I_i^{T}\}$ are deformed versions of the reference image, simulated by applying affine transformations such as scaling, rotation, and translation. We split the dataset into $1000/200/200$ sequences for training, validation, and testing. \\
\noindent \textbf{Cardiac MRI Series.} We include $600$ cine CMR videos collected from sixty subjects. Each video sequence covers half of the cardiac motion cycle (with $T+1=7$), spanning from the peak stretching phase to the peak contracting phase. We crop the original image to the size of $64 \times 64$, focusing on the structure of left ventricles. The LV myocardium of all video sequences are manually annotated by clinical experts. We randomly choose $400$ sequences from $40$ subjects for training, $100$ sequences from $10$ subjects for validation, and the rest for testing.

\subsection{Experimental Design}
We compare the performance of TLRN with four state-of-the-art deep learning-based diffeomorphic registration algorithms in the context of both SVF~\cite{arsigny2006log} and LDDMM~\cite{beg2005computing}. These approaches include Voxelmorph (VM)~\cite{dalca2019unsupervised}, TransMorph (TM)~\cite{chen2022transmorph}, SVF-R2Net~\cite{joshi2022diffeomorphic}, and Lagomorph (LM)~\cite{hinkle2018diffeomorphic}. All methods are trained on the same dataset with their best performance reported. 

Note that we consider the comparison with VM~\cite{dalca2019unsupervised} as an ablation study for the proposed temporal latent residual learning block. Both methods use U-Net as the network backbone with the same number of convolutional layers. We maintain consistent network parameters, including regularity weights on velocity fields and the timesteps of integration, to ensure a fair comparison.

\paragraph*{\bf Evaluation Metric.} We first evaluate the registration accuracy of TLRN on synthetic dataset by computing the mean squared error (MSE) between deformed sequence images and target sequence images over the time and comparing the results with all baselines. We report the percentage of negative Jacobian determinants of predicted transformation fields from our method vs. all baselines.

We assess the registration accuracy on CMR videos by performing registration-based segmentation of the left ventricular (LV) myocardium. The propagated segmentations by deforming manually segmented myocardium on the reference image using predicted transformation fields from all methods are then compared. To evaluate volume overlap between the propagated segmentation $A$ and the manual segmentation $B$, we compute the dice similarity coefficient (DSC)~\cite{dice1945measures} by DSC$(A, B) = 2(|A| \cap |B|)/(|A| + |B|)$, where $\cap$ denotes an intersection of two regions. Additionally, we measure the maximum discrepancy between boundaries of the propagated segmentation and the manual segmentation by Hausdorff distance (HD)~\cite{huttenlocher1993comparing}. Given two sets of boundary points $X \in A$ and $Y \in B$ for LV myocardium, we compute the distance by HD$(X,Y) = max \{h(X,Y), h(Y,X) \}$, where $h(X,Y) = \mathop{max}_{x_i \in X} \mathop{min}_{y_i \in Y}  \| x_i-y_i \|$, and vice versa. 
\subsection{Results}
\begin{figure*}[!hb]
\centering
\includegraphics[width=0.95\textwidth] {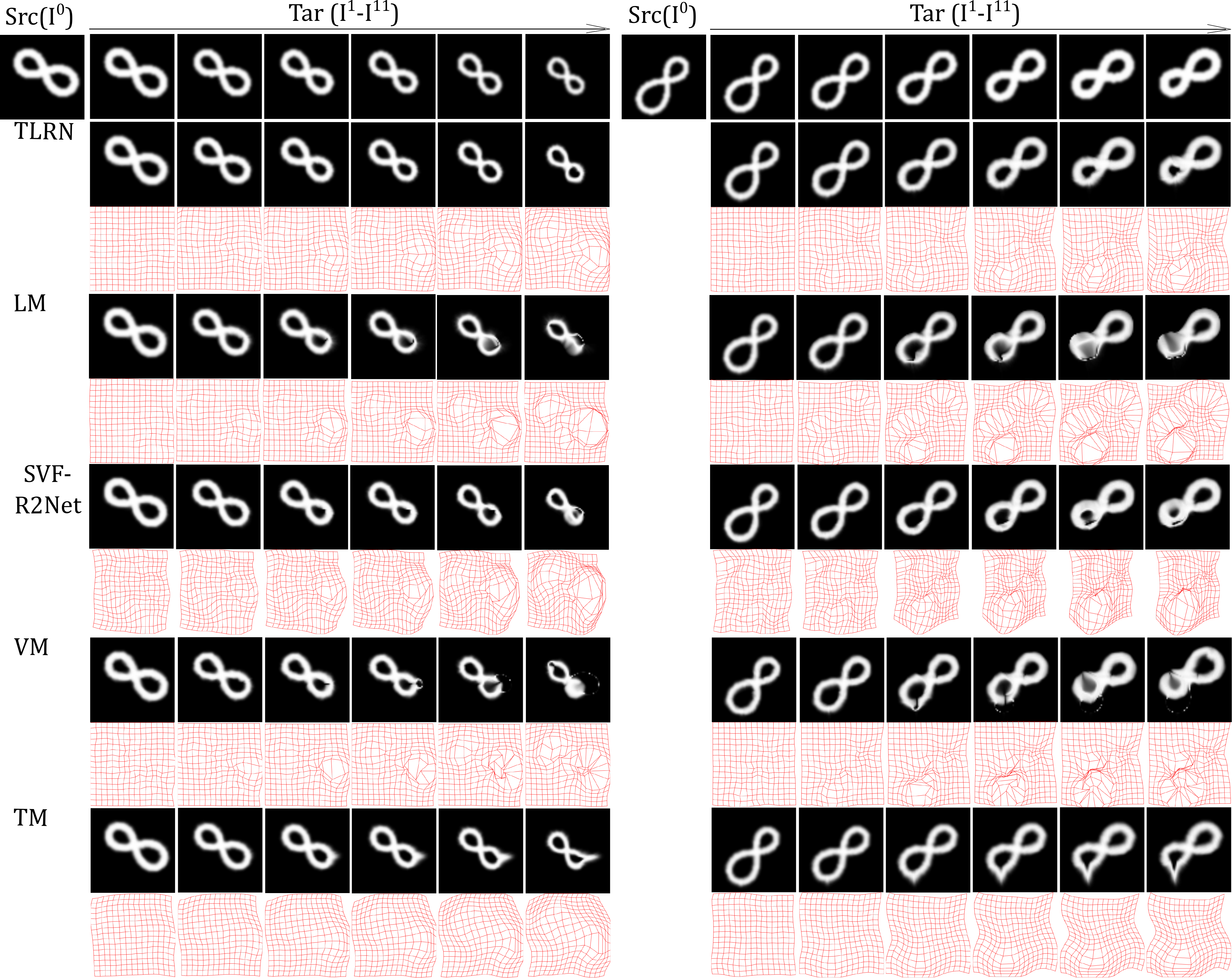}
     \caption{Left to right: examples of deformed reference/source image across time. Top to bottom: a comparison of deformed images and transformation fields predicted from our model TRLN and baselines.}
\label{fig:synthetic}
\end{figure*}

Fig.~\ref{fig:synthetic} visualizes the predicted time-series of both deformed images and deformation fields (from $I^0$ to the remaining frames $I^1 \sim I^T$) from all methods. It shows that our method, TRNL, consistently achieves improved quality of deformed images with well regularized transformation fields along the time points. Moreover,  our model obtains the lowest percentage of negative Jacobian determinants (\underline{TLRN: $0.175\%\pm 0.373\%$}; LM: $0.474\% \pm 0.969\%$, SVF-R2Net: $1.607\% \pm 1.917\%$, VM: $0.398\% \pm 0.591\%$, and TM: $0.176\% \pm 0.471\%$), which indicates better quality and more realistic registration results.

\begin{figure*}[b!]
\centering
\includegraphics[width=1.0\textwidth] {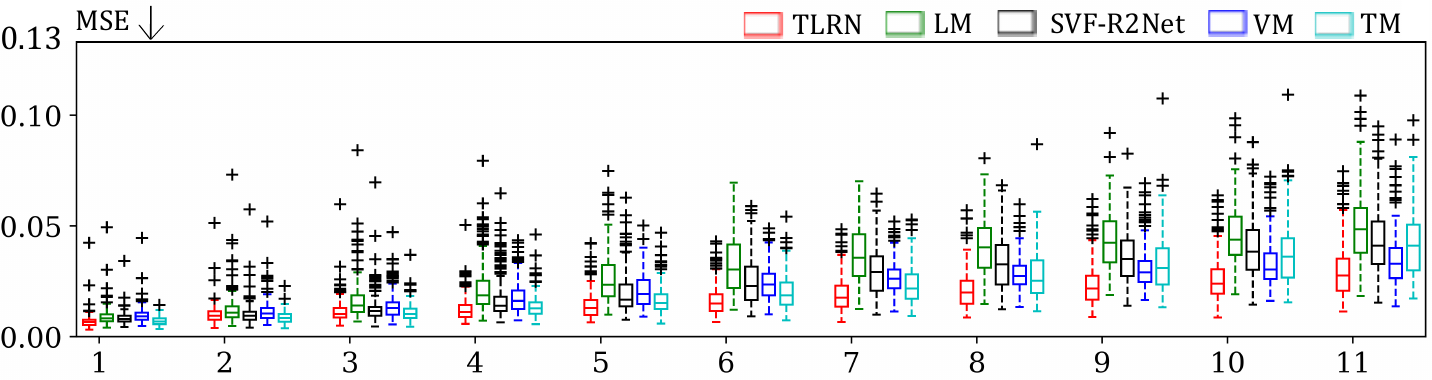}
     \caption{A comparison of MSE between deformed and target time-series images.}
\label{fig:mse}
\end{figure*}

Fig.~\ref{fig:mse} compares MSE between the deformed images and the target images across all methods over multiple time steps. It shows that TLRN achieves substantially lower MSE compared to all baselines. The improved registration accuracy and smoothness of deformation fields (especially on later time frames when large deformations occur) indicate that our model effectively leverages the spatiotemporal continuity within the cardiac motion sequence. 

\begin{figure*}[!h]
\centering
\includegraphics[width=0.95\textwidth] {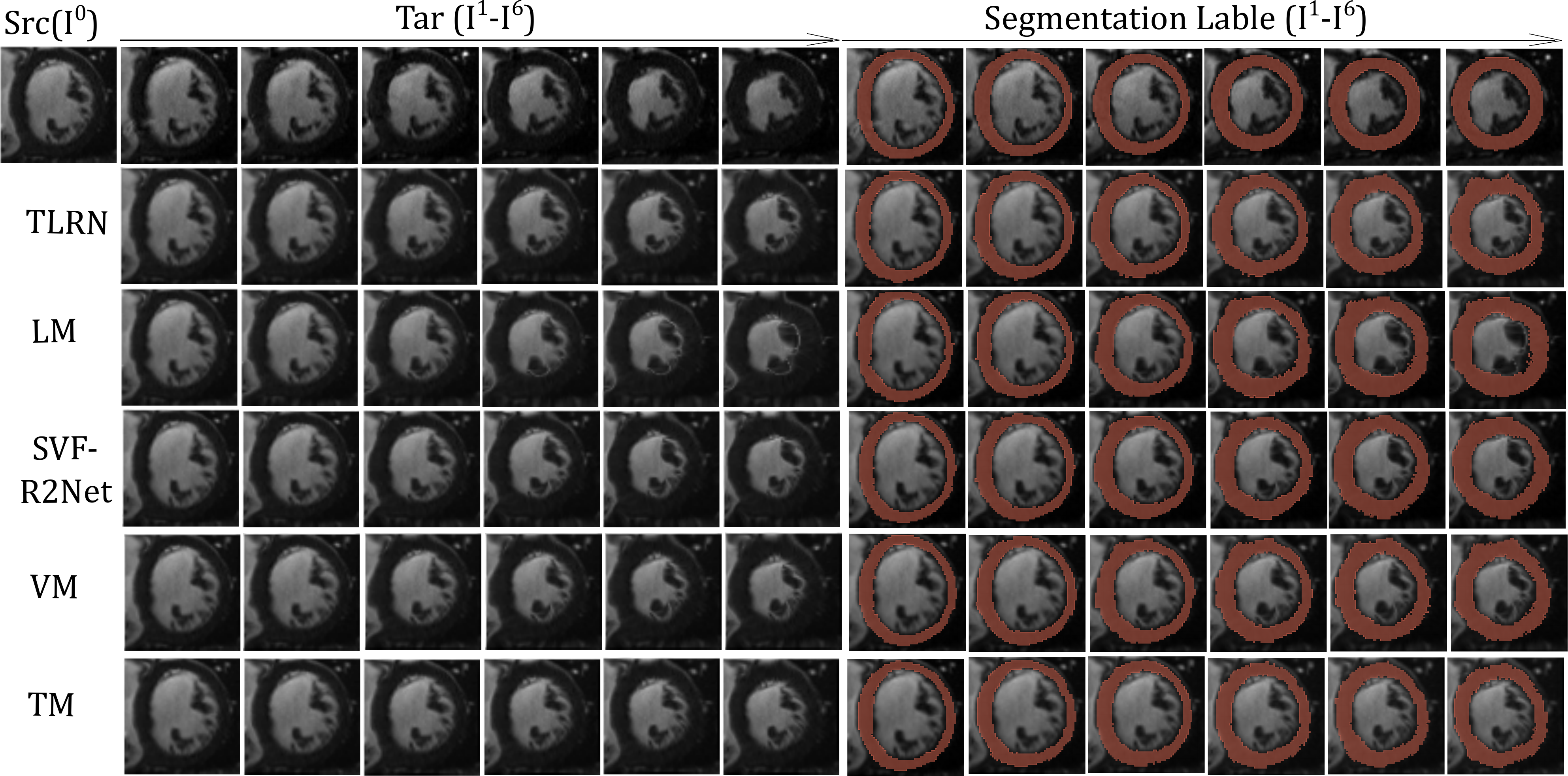}
     \caption{Left to right: examples of CMR image videos and overlaid LV myocardium segmentation maps. Top to bottom: a comparison of manually delineated segmentation lables vs. propagated segmentation from all methods.}
\label{fig:labels}
\end{figure*}

\begin{figure*}[t!]
\centering
\includegraphics[width=0.95\textwidth] {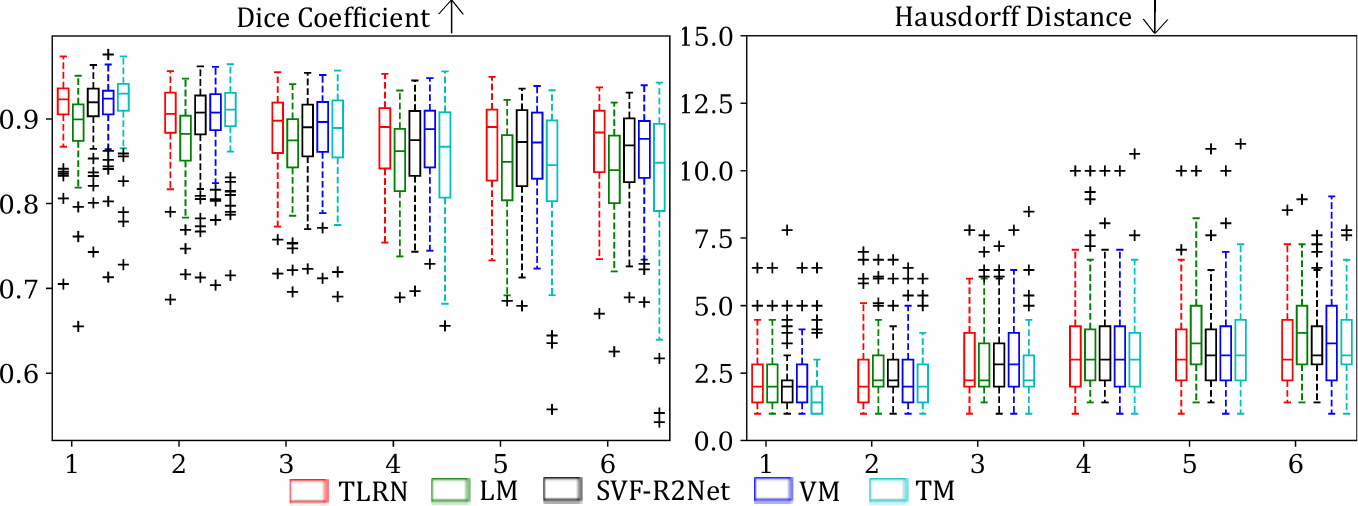}
     \caption{Left to right: a comparison of dice vs. Hausdorff distance score on predicted LV myocardium segmentation labels from all methods over the time frames $\tau (1 \sim 6)$.}
\label{fig:hausdorff}
\end{figure*}
The left panel of Fig.~\ref{fig:labels} presents visual examples of the reference/source frame, target sequence, and deformed sequence over all methods. The right panel of Fig.~\ref{fig:labels} displays a comparison between the manually labeled segmentation on LV myocardium and propagated segmentation labels deformed by deformations fields predicted from all methods.

Fig.~\ref{fig:hausdorff} displays a quantitative comparison between the manually delineated vs. propagated segmentation labels deformed by deformations fields predicted from all methods. The left panel reports the statistics of dice scores (the higher the better), and the right panel reports the statistics of Hausdorff distance (the lower the better) on the LV myocardium. All metrics are computed throughout the cardiac motion cycle, from the peak contraction phase to the peak stretching phase. It shows that TLRN is able to produce superior quality of final dice and HD scores in comparison with other baselines.

\section{Conclusions \& Discussion}
This paper presents a novel temporal latent residual network, TLRN, to predict a sequence of deformation fields in time-series image registration. To the best of our knowledge, our model is the first to develop a spatio-temporal network with residual blocks in the latent deformation space, parameterized by velocity fields. The learned residual functions over time are well utilized to effectively adjust the continuous deformation features from current and previous time frames. Experimental results on both synthetic sequences and real-world cine CMR videos show that the proposed TLRN is able to achieve an improved registration accuracy with better regularized deformation fields. 

\begin{credits}
\subsubsection{\ackname} This work was supported by NSF CAREER Grant 2239977 and NIH 1R21EB032597.

\subsubsection{\discintname}
The authors have no competing interests to declare that are relevant to the content of this article.
\end{credits}

\bibliographystyle{splncs04}
\bibliography{Paper-3610}

\end{document}